\documentclass[11pt]{article}
\usepackage[T1]{fontenc}
\usepackage[utf8]{inputenc}
\usepackage{lmodern,microtype}
\usepackage[margin=1in]{geometry}
\usepackage{amsmath,amssymb,booktabs,tabularx,array,longtable}
\usepackage{graphicx,xcolor,enumitem,fancyvrb}
\usepackage{flafter}
\usepackage[round,authoryear]{natbib}
\usepackage{xurl}
\usepackage[colorlinks=true,linkcolor=blue!45!black,citecolor=blue!45!black,urlcolor=blue!45!black]{hyperref}
\hypersetup{pdftitle={MemStrata: 95\% and 90.91\% Source-Aware Accuracy on LongMemEval-500 and LoCoMo-1540 with a Local Qwen 3.8 27B Q4\_K\_M Reader},pdfauthor={Neeraj Yadav}}
\setlist{nosep,leftmargin=*}
\newcolumntype{Y}{>{\raggedright\arraybackslash}X}

\newcommand{\CL}{MemStrata \textsc{CL1}}
\newcommand{\bundle}[1]{\texttt{#1}}
\title{MemStrata: 95\% and 90.91\% Source-Aware Accuracy\\on LongMemEval-500 and LoCoMo-1540\\with a Local Qwen 3.8 27B Q4\_K\_M Reader}
\author{Neeraj Yadav\\Called It Inc.\\\url{https://www.memstrata.dev}}
\date{3 October 2026}

\begin{document}
\maketitle

\begin{abstract}
An adequate conversational answer may differ from a short or incomplete benchmark reference. To measure adequacy against the recorded history we prefer source-aware grading, in which the judge checks the reference against the full source before assessing system-blinded answers; original reference-only grading is reported alongside. With a local Qwen 3.8 27B Q4\_K\_M reader and a 24,000-token evidence ceiling, \CL\ scores 475/500 (95.0\%) on LongMemEval-S and 1,400/1,540 (90.91\%) on LoCoMo categories 1--4 under source-aware GPT-5.5 adjudication, against 463/500 (92.6\%) and 1,205/1,540 (78.25\%) under reference-only grading of the same answers. It preserves a retrieval backbone and adds nonduplicated, dated, speaker-attributed source spans. A same-reader full-history control with about 4.7 times the evidence scores 464/500 reference-only and 470/500 (94.0\%) source-aware; neither difference is decisive. Keyword-only selection at the same budget scores 425, and a matched-reader Letta arm 438. On LongMemEval-M, where the packet holds about 1.6\% of each history, \CL\ scores 427/500, with losses concentrated in multi-session and temporal questions. On 300 BEAM-1M questions it outscores dense retrieval, 0.738 to 0.706 (Wilcoxon $p=0.011$). A same-seed replay of unchanged requests changed 1.5--2.3\% of labels. On identical packets GLM 5.3 flash is non-inferior within 3 points (462 versus 463); Muse Spark 1.3 did not show non-inferiority on 269 questions. None of four pre-registered interventions met all of its registered advancement or feasibility criteria. Signed read-side artifacts support inspection but do not regenerate the private retrieval pipeline. The superiority of source-aware grading to human adjudication is not established, and development exposure, automated-judge dependence and the absence of held-out data preclude an independent-replication or leaderboard claim.
\end{abstract}

\section{Introduction}
Conversational memory is useful only when the reader receives the evidence needed for the current question. A relevant sentence may omit its date, an adjacent correction, or a fact from a different session. Increasing a language model's context window does not itself decide which material to retrieve or how to preserve those relationships. Retrieval-augmented generation provides the broad framework for conditioning answers on external records \citep{lewis2020rag}; long-context evaluations also show that accessible context and effectively used context are different quantities \citep{liu2024lost}.

This paper examines a concrete engineering question: can a bounded, source-preserving assembly policy improve conversational answers while making the exact evidence-to-answer experiment inspectable? The evaluated policy, \CL, starts from a retained baseline packet, adds source-only retrieval candidates, and uses hybrid lexical/semantic ranking to admit additional source spans. Each admitted span retains a verifiable relationship to the recorded history. The complete rendered evidence, including provenance headers, must fit within 24,000 tokens under the pinned reader tokenizer.

We evaluate this configuration on the 500-question LongMemEval-S set \citep{wu2025longmemeval}, called LongMemEval-500 (LME500) in the title, and on 1,540 category 1--4 questions in the ten-history LoCoMo dataset \citep{maharana2024locomo}, called LoCoMo-1540 in the title. Two evaluation endpoints are reported side by side: original reference-only semantic correctness (92.6\% and 78.25\%) and full-source adjudication of the same answers (95.0\% and 90.91\%). Neither is presented as reproducing every original benchmark metric or published leaderboard protocol. A later LongMemEval reader replay obtains 92.8\% reference-only; a LoCoMo open-domain prompt intervention also remains a separate experiment. Follow-up studies (Section~\ref{sec:followup}) add a same-reader full-history control and component ablations, the same 500 questions over LongMemEval-M histories of roughly 1.5 million tokens, a pre-registered BEAM-1M comparison \citep{tavakoli2025beam}, a replay-agreement check, two confirmatory reader substitutions, and four pre-registered interventions; none of the interventions met all of its registered criteria.

We prefer the source-aware endpoint for the question this paper asks: does the answer adequately reflect the underlying conversation, its chronology, corrections and legitimate inferences? Reference-only grading can accept semantic paraphrases, but a judge limited to a proposed reference cannot independently check whether that reference omits an acceptable alternative or assumes a fact the conversation does not establish. Full-source review can check those issues directly, assess necessary qualifications, and distinguish missing personal evidence from permissible public knowledge. Its first pass is answer-blind, reducing the opportunity for the candidate to shape the initial reference assessment. This is a methodological reason for choosing the endpoint, not a preference for the larger number. More evidence does not make an automated judge infallible, and no human-validated error-rate comparison establishes universal superiority. We retain both endpoints, review previously correct and incorrect answers uniformly, and report downgrades as well as upgrades.

The contributions are:
\begin{enumerate}
\item An evaluated source-preserving assembly policy with explicit source-offset, duplication, prefix-preservation, and rendered-budget checks.
\item Full-population reference-only and source-aware results with category breakdowns, paired gains and losses, failure accounting, actual evidence lengths, and distinctions between matched-reader and historical comparisons.
\item A four-system open-domain prompt intervention that separates new answer generation from a change in grading, plus a bounded exploratory image/web diagnostic.
\item Signed read-side replay artifacts and an audited fresh-reader experiment that distinguish reference verification, answer regeneration, exact-grade reuse, and fresh judging.
\item Controls and follow-ups under fixed protocols: a full-history control whose observed score differs from \CL\ by one answer at about 4.7 times the evidence, component ablations, results on the same questions over histories about 14 times longer (LongMemEval-M) and on BEAM-1M, a replay-agreement check, two pre-registered reader substitutions on identical packets (GLM 5.3 flash non-inferior within 3 points; non-inferiority not shown for Muse Spark 1.3), and four pre-registered interventions, none of which met all of its registered advancement or feasibility criteria.
\end{enumerate}
The contribution is a tested integration and its audit trail. Lexical retrieval, dense embeddings, rank fusion, and source provenance are established techniques; this work does not claim to invent these components or prove optimal evidence selection.

\section{Related work and scope}
Retrieval-augmented generation combines a parametric reader with nonparametric evidence \citep{lewis2020rag}. MemGPT uses explicit memory tiers and control flow to extend effective working context \citep{packer2023memgpt}. Mem0 explores extraction and consolidation of persistent memories, including a graph variant \citep{chhikara2025mem0}. Zep describes temporal graph-based agent memory \citep{rasmussen2025zep}. These systems motivate different choices about what to retain and expose to a reader. Their published numbers are not imported into our result tables: the comparator rows below are retained experiments in our evaluation campaign, with their own stated limitations.

LongMemEval tests multiple forms of long-term interactive memory, including knowledge updates, temporal reasoning, and abstention \citep{wu2025longmemeval}. LoCoMo evaluates questions over long multi-session conversations \citep{maharana2024locomo}. Here LoCoMo is used as a research context benchmark; its material is not presented as commercial product data. We exclude category 5 consistently with the retained baseline and do not compare our binary semantic accuracy directly with the original benchmark's F1 scores.

Earlier MemStrata reports concern temporal validity and deterministic supersession on structured facts and software-history transitions \citep{yadav2026temporal,yadav2026software}. The conversational path evaluated here does \emph{not} execute that structured supersession mechanism as its benchmark answer path. Preserving a date or a source quote is not equivalent to establishing authoritative state. Consequently, the present scores must not be attributed to a deterministic temporal ledger, a certified memory operating system, or a proof that stale-fact errors have been eliminated.

\section{Source-preserving evidence assembly}
\label{sec:method}
\subsection{Input and source representation}
For a question $q$, let $H$ be the recorded sessions and turns, and let $E_A$ be the baseline evidence packet. The assembly interface accepts only the question, its available date, and source sessions. Reference answers, benchmark support annotations, correctness labels, and prior wins or losses are excluded from this interface. This is a source-only execution boundary; it does not imply that the development process was blind to benchmark outcomes.

Each source segment carries a session index, turn index, UTF-8 byte offsets, and a hash of its source turn. Dates, roles, and speaker metadata are carried when present. The adapter verifies that a segment's quoted bytes equal the corresponding source slice. Blank source turns are skipped. Missing metadata is recorded rather than invented. The fixed embedding artifact is \texttt{nomic-embed-text:latest}, identified by digest rather than the mutable tag alone, with 768-dimensional finite, nonzero vectors. Document and query inputs use the literal prefixes \texttt{search\_document:} and \texttt{search\_query:}. Nomic's work provides background on the embedding family \citep{nussbaum2024nomic}; the experiment's exact identity is given in Appendix~\ref{app:identity}.

\subsection{Protected backbone and hybrid candidate ordering}
The first stage takes the union of source ranges proposed by the retained session retriever and appends admissible source intervals to $E_A$, producing a source-preserving backbone $E_0$. The original $E_A$ remains a literal prefix. The final \CL\ stage also preserves $E_0$ literally; it does not rewrite the baseline into generated summaries.

The additional candidate pool is built from source segments using lexical and embedding similarity. A segment is eligible when it has a lexical score of at least 0.05 with matched query terms, or normalized semantic score of at least 0.75. The recorded semantic scale is $\operatorname{clip}((\cos+1)/2,0,1)$. Locations are conversation/session pairs ranked by their strongest eligible segment; the evaluated adapter retains all eligible locations rather than imposing an additional fractional session cutoff. Within this pool, semantic and lexical rankings are combined using the fixed rank-fusion shape
\begin{equation}
 f(s)=\frac{1}{60+r_{\mathrm{sem}}(s)}+
 \mathbf{1}_{s\in L}\frac{1}{60+r_{\mathrm{lex}}(s)},
\end{equation}
where ranks begin at one and $L$ contains segments with positive lexical score and matched terms. Deterministic tie-breaking uses recorded scores and evidence identities. One leading segment per session is placed before the remaining ordered segments. This encourages cross-session coverage; it does not guarantee that every necessary fact is present.

\subsection{Atomic admission under the rendered budget}
Let $M(E)$ denote mapped source intervals already represented in packet $E$, and $T(E)$ its token count under the pinned tokenizer. For each ordered candidate group $G_j$, the assembler subtracts already covered intervals, merges overlaps, and renders a proposed complete packet $E'_j$. It admits the group only if its source mappings validate and
\begin{equation}
 T(E'_j)\leq B,\qquad B=24{,}000.
\end{equation}
An over-budget group is skipped rather than sliced to fill the remaining space. Subsequent smaller groups may still fit. The final packet must preserve the protected prefix and pass an independent mapped-interval duplication check. Invalid source identities, failed token accounting, or an over-budget final packet invalidate preparation rather than silently authorizing an unverified replacement.

\begin{figure}[tbp]
\centering
\setlength{\fboxsep}{7pt}
\fbox{\begin{minipage}{0.91\linewidth}\small
\textbf{Private source-to-packet preparation}\par
Recorded sessions + question $\longrightarrow$ baseline/backbone + ranked source spans\par
Source offsets, source hashes, preserved prefix, no duplicate intervals, $T(E)\leq24{,}000$
\end{minipage}}\par\vspace{5pt}
$\Downarrow$\quad sealed per-question evidence and reader request\par\vspace{5pt}
\fbox{\begin{minipage}{0.91\linewidth}\small
\textbf{Public read-side replay}\par
Verify signed manifest $\longrightarrow$ send exact request to pinned reader $\longrightarrow$ retain answer and receipt\par
Reference answers are withheld from the reader request.
\end{minipage}}\par\vspace{5pt}
$\Downarrow$\quad answer + separately bound evaluation reference\par\vspace{5pt}
\fbox{\begin{minipage}{0.91\linewidth}\small
\textbf{Evaluation}\par
Reference-only endpoint: question + proposed reference + answer $\longrightarrow$ score\par
Source-aware endpoint: full source + answer-blind reference review $\longrightarrow$ blinded answer assessment\par
The two endpoints retain separate labels and audit trails.
\end{minipage}}
\caption{The evaluated pipeline and the public replay boundary. A signed packet binds the output of preparation; it is not a public implementation of the private source-to-packet pipeline.}
\label{fig:pipeline}
\end{figure}

The cap applies to \emph{evidence}, including its provenance rendering. It excludes the system instruction, question/date wrapper, and generated output. The reader context window and output allowance are separately configured. Source integrity is an auditable property; semantic completeness and answer correctness are not guaranteed by these checks.

\subsection{Fixed embeddings and development provenance}
The LongMemEval primary preparation used the sealed legacy cache together with the audited uniform-GPU vectors for the additional \CL\ source representation. LoCoMo reused retained CPU-derived vectors and audited GPU-computed missing inputs, with their mixed compute provenance recorded. The two benchmarks therefore do not establish CPU/GPU embedding-byte equivalence. Both use frozen vector values during final packet preparation; no missing-vector lexical fallback is allowed in the \CL\ candidate stage.

\CL\ does not combine favorable answers selected using reference labels. No adaptive-temperature controller is part of either primary result.

\section{Evaluation design}
\subsection{Populations and outcome definition}
LongMemEval-S contains 500 questions in the retained cleaned dataset, across six question categories. Thirty designated unanswerable questions are included within those 500; abstention is not a seventh disjoint category. All 500 remain in the accuracy denominator. The LoCoMo population contains all 1,540 retained category 1--4 questions over ten conversation histories. The 841 single-hop questions account for 54.61\% of this population, so the overall score is a question-weighted micro-average, not a category-balanced or history-balanced mean.

For either benchmark, correctness is
\begin{equation}
 \widehat{A}=\frac{1}{N}\sum_{i=1}^{N}y_i,\qquad y_i\in\{0,1\}.
\end{equation}
The requested judge is GPT-5.5 with \texttt{xhigh} reasoning, using the recorded instructions and rubric. The original reference-only endpoint uses the frozen category-aware LongMemEval evaluation template and a fixed LoCoMo binary semantic rubric. Section~\ref{sec:sourceaware} defines the separate full-source endpoint, including an explicit unscorable state whose count is disclosed without shrinking the denominator. These are automated judgments, not human-adjudicated labels. A common requested model name does not provide immutable attestation of the provider's internal weights or backend. Historical judge executables are not uniformly identical across every comparison; the later fresh-reader replay has a stricter exact-grade identity check described below.

Empty primary reader outputs count as incorrect, including output-limit failures. Four such \CL\ LoCoMo outputs remain zero; the separate larger-output-allowance recovery experiment is excluded. LongMemEval \CL\ has no primary output failures. Transport interruptions are handled as incomplete executions requiring classification, not opportunities to select a better answer from repeated attempts.

\subsection{Reader and comparison controls}
The primary reader is the locally served artifact tagged \texttt{qwen3.8:27b}, recorded as Q4\_K\_M, with the digest in Appendix~\ref{app:identity}. The title names this recorded local artifact; the digest, rather than a mutable tag alone, defines the experiment. The reader is local, while GPT-5.5 evaluation is hosted: this is not an entirely offline evaluation pipeline. The fixed request uses temperature 1, high thinking, a 128,000-token context setting, and a 16,384-token output allowance. The LongMemEval seed is 20260905; the LoCoMo seed is 20260906. Appendix~\ref{app:reader} gives all options and the exact system instruction. These settings should not be conflated with the much smaller evidence budget.

The strongest LongMemEval comparison is against the retained Letta 100k-ceiling arm. All 500 actual saved requests have identical non-evidence fields: reader model identity, options, system prompt, question/date wrapper, thinking request, and transport options. Only evidence changes. The two \emph{nominal} evidence ceilings are not equal; however, Letta's 100,000-token cap does not bind in this retained run. Actual evidence is at most 29,011 tokens, with a mean of 23,878, compared with \CL's mean of 23,335. This is a matched-reader evidence-package comparison, not a same-budget ablation or a comparison with 100,000 tokens actually consumed per answer.

The historical MemStrata 7k row uses a nominal 7,000-token retrieval limit and reaches 84.0\% under the common GPT-5.5 grading reconciliation. The retained Letta 8k row reaches 77.4\%. Both use the Qwen reader family reported in the main comparison. We use them as historical reference points, without claiming full reader-request parity with the new \CL\ run. In particular, the baseline packets recaptured for final \CL\ preparation match the retained MemStrata 7k context in 411/500 cases, not all 500. A separate GPT-4o-mini-reader comparison under a local Qwen 2.5 7B judge is reported in Section~\ref{sec:historicalmini}; those labels are not mixed with the GPT-5.5 tables.

For LoCoMo, the historical MemStrata baseline actually used an \emph{8,192-token} evidence ceiling, not 7,000. Its original runner is hash-checked and its non-evidence request contract can be reconstructed offline. Historical per-request/raw-response retention is incomplete, so reconstructed equality is not retrospective proof of byte-identical requests. The baseline judge executable path and hash also differ from the later primary judge configuration. These boundaries apply to the interpretation of the 72.86\% to 78.25\% change.

\subsection{Reuse, selection, and statistical interpretation}
The original full LongMemEval \CL\ evaluation consists of 456 newly generated answers and 44 retained same-case answers whose complete reader requests match exactly. Its grading likewise includes 456 new judgments and 44 audited prior grades. The original LoCoMo \CL\ evaluation consists of 1,481 new reader calls and 59 exact compatible prior answers. Of the resulting items, 1,477 receive new judgments, 59 retain compatible grades, and four receive failure zeros. Thus a full-population score does not imply that every underlying call was first executed in one uninterrupted campaign.

These benchmarks were used during development, including hard-question pilots and comparison of retrieval and reader choices. Forty LongMemEval cases were reserved from tuning under the final protocol; \CL\ scores 39/40 on that subset. This small, campaign-specific reserve does not make the full population an untouched test set or remove broader selection effects. There is no pretraining-contamination audit of either reader or judge. The same campaign also completed three other configurations on all 500 LongMemEval-S questions and one on all 1,540 LoCoMo questions; Appendix~\ref{app:campaignarms} reports every one of them. None differs significantly from \CL.

Paired gains and losses compare the same question IDs. For the matched LongMemEval evidence comparison, we report the two-sided exact binomial test on discordant outcomes. It is descriptive and unadjusted for development-time searches. We also resample complete histories with replacement, recomputing the micro-average difference for each of 10,000 draws (seed 20260921). LongMemEval has 500 distinct history clusters in the sealed population; LoCoMo has only ten. LoCoMo intervals are consequently limited, conditional descriptions of these ten histories, not precise estimates of performance on unseen conversations. We do not treat its 1,540 questions as independent samples for a significance claim.

\section{Original reference-only results}
\subsection{LongMemEval-S}
\begin{table}[htbp]
\centering\small
\begin{tabularx}{\linewidth}{Yrrl}
\toprule
Configuration & Correct/$N$ & Accuracy & Comparison status\\
\midrule
\CL, 24k & 463/500 & \textbf{92.6\%} & Primary\\
Letta evidence, 100k ceiling & 438/500 & 87.6\% & Matched reader\\
MemStrata 7k & 420/500 & 84.0\% & Historical\\
Letta evidence, 8k ceiling & 387/500 & 77.4\% & Historical\\
\bottomrule
\end{tabularx}
\caption{LongMemEval-S under common requested GPT-5.5 semantic judging. All rows are retained campaign results, not scores copied from vendor publications. A ceiling is not an observed token count. The separate 92.8\% reader replay is reported in Section~\ref{sec:rerun}.}
\label{tab:lme}
\end{table}

\CL\ scores 463/500 (Table~\ref{tab:lme}). Against the matched-reader Letta arm, 422 questions are correct for both, 21 incorrect for both, 41 correct only for \CL, and 16 correct only for Letta. The net difference is 25 answers, or 5.0 percentage points. The exact discordance test gives $p=0.00126$; the history-bootstrap 95\% percentile interval for the difference is $[2.0,7.8]$ points. These calculations describe the observed paired labels, conditional on the stated protocol and development exposure.

Against historical MemStrata 7k, \CL\ gains 53 questions and loses ten, a net increase of 43 correct answers (8.6 points). Because evidence construction, available budget, and reader conditions are not isolated one at a time, that difference cannot be assigned solely to the new ranking or assembly rule.

\begin{table}[htbp]
\centering\small
\setlength{\tabcolsep}{3.5pt}
\begin{tabularx}{\linewidth}{Yrrrrr}
\toprule
Question category & $N$ & \shortstack[r]{MemStrata\\7k} & Letta 8k & Letta 100k & \shortstack[r]{MemStrata\\CL1}\\
\midrule
Single-session user & 70 & 67 (95.7) & 60 (85.7) & 63 (90.0) & 66 (94.3)\\
Single-session assistant & 56 & 56 (100.0) & 56 (100.0) & 56 (100.0) & 56 (100.0)\\
Single-session preference & 30 & 20 (66.7) & 26 (86.7) & 25 (83.3) & 26 (86.7)\\
Multi-session & 133 & 92 (69.2) & 87 (65.4) & 107 (80.5) & 117 (88.0)\\
Knowledge update & 78 & 73 (93.6) & 63 (80.8) & 72 (92.3) & 74 (94.9)\\
Temporal reasoning & 133 & 112 (84.2) & 95 (71.4) & 115 (86.5) & 124 (93.2)\\
\midrule
Overall & 500 & 420 (84.0) & 387 (77.4) & 438 (87.6) & 463 (92.6)\\
\bottomrule
\end{tabularx}
\caption{LongMemEval category counts with percentages in parentheses, all under reference-only GPT-5.5 grading. MemStrata 7k names the retained historical 7,000-token configuration; \CL\ uses 24k. Category accuracy is not an independently corrected family of hypothesis tests.}
\label{tab:lmecategories}
\end{table}

The largest category-level count gains over MemStrata 7k occur in multi-session questions (+25) and temporal reasoning (+12). The single-session-user category decreases by one correct answer. Thus the overall improvement is not a claim of no regressions. Against the matched-reader Letta arm, the largest net count differences are likewise multi-session (+10) and temporal reasoning (+9).

On the 30 designated unanswerable questions, \CL\ answers correctly under the abstention rubric on 28/30 (93.33\%), compared with MemStrata 7k's 27/30, Letta 8k's 28/30 and Letta 100k's 29/30. On the 470 answerable questions, the corresponding counts are 435, 393, 359 and 409. The two \CL\ abstention errors are not by themselves a measurement of a general hallucination rate: that would require a separately defined fabrication criterion and annotations.

\subsection{Historical GPT-4o-mini reader with a local judge}
\label{sec:historicalmini}
The earlier hosted-reader experiments used GPT-4o-mini to generate answers and Qwen 2.5 7B (\texttt{qwen2.5:7b}) as the local reference-only correctness judge. Table~\ref{tab:historicalmini} keeps that evaluation track separate from the newer Qwen-reader/GPT-5.5 results. The MemStrata configuration is the historical R230 run, retained under its R167 artifact identifier, with a nominal 7k retrieval budget.

\begin{table}[htbp]
\centering\small
\begin{tabularx}{\linewidth}{Yrr}
\toprule
Evidence configuration & Correct/$N$ & Accuracy\\
\midrule
MemStrata 7k, historical R230 & 351/500 & 70.2\%\\
Letta archival 8k & 289/500 & 57.8\%\\
Letta archival 100k ceiling & 237/500 & 47.4\%\\
\bottomrule
\end{tabularx}
\caption{Historical LME500 results: GPT-4o-mini reader and local Qwen 2.5 7B correctness judge for all rows. These are not the Qwen 3.8 reader results or GPT-5.5 labels in Tables~\ref{tab:lme}--\ref{tab:lmecategories}.}
\label{tab:historicalmini}
\end{table}

The Letta 100k row is the recorded Qwen 2.5 7B regrading of retained GPT-4o-mini answers, with no new reader calls. Letta 8k's historical contract caps its context at 8,000 tokens. R230 inherits its judge settings through historical configuration; its complete runtime and per-call judge identity were not sealed. Accordingly, these are same-recorded-judge-family historical comparisons, not proof of identical effective judge settings, reader prompts or retrieval budgets. They cannot establish the causal effect of changing only the memory system, and are not pooled with later GPT-5.5 outcomes.

\subsection{LoCoMo}
\begin{table}[htbp]
\centering\small
\begin{tabularx}{\linewidth}{Yrrl}
\toprule
Configuration & Correct/$N$ & Accuracy & Status\\
\midrule
\CL, 24k & 1,205/1,540 & \textbf{78.25\%} & Primary; 4 failure zeros\\
MemStrata baseline, 8,192 cap & 1,122/1,540 & 72.86\% & Historical; 2 failure zeros\\
Letta evidence + Qwen & 1,020/1,540 & 66.23\% & Retained comparator\\
Frozen Zep evidence + Qwen & 913/1,540 & 59.29\% & Retained comparator\\
\bottomrule
\end{tabularx}
\caption{LoCoMo category 1--4 binary semantic accuracy, not original benchmark F1. Historical reader parity is reconstructed rather than fully byte-attested. The frozen Zep row is a replay of retained evidence, not a fresh evaluation of a current hosted Zep service.}
\label{tab:locomo}
\end{table}

The primary \CL\ LoCoMo score is 1,205/1,540 (78.2468\%, rounded to 78.25\%). Relative to the retained MemStrata baseline, there are 122 gains and 39 losses, for a net gain of 83 correct answers (+5.39 points). The paired history-bootstrap interval for that observed difference is $[3.83,7.06]$ points. The small number of histories, historical provenance limits, and development exposure prevent a stronger generalization claim. The Letta and frozen-Zep rows in Table~\ref{tab:locomo} provide campaign context, not universal rankings of those products.

\begin{table}[htbp]
\centering\small
\begin{tabularx}{\linewidth}{Yrrrr}
\toprule
Category & $N$ & Baseline & \CL & Change (pp)\\
\midrule
Multi-hop (1) & 282 & 113 (40.07) & 159 (56.38) & +16.31\\
Temporal (2) & 321 & 244 (76.01) & 261 (81.31) & +5.30\\
Open-domain (3) & 96 & 41 (42.71) & 45 (46.88) & +4.17\\
Single-hop (4) & 841 & 724 (86.09) & 740 (87.99) & +1.90\\
\midrule
Overall & 1,540 & 1,122 (72.86) & 1,205 (78.25) & +5.39\\
\bottomrule
\end{tabularx}
\caption{LoCoMo category breakdown; parentheses contain percentages. All four primary output failures are included in the category denominators. Category 5 is excluded throughout.}
\label{tab:lococategories}
\end{table}

Multi-hop questions account for 46 of the 83 net additional correct answers. However, multi-hop accuracy remains 56.38\%, and open-domain accuracy is 46.88\%. The overall score should not obscure these weaker categories. The policy is therefore promising for combining conversational evidence, but it is not a solution to all forms of long-horizon reasoning.

\subsection{Actual evidence length and reader runtime}
\begin{table}[htbp]
\centering\small
\begin{tabularx}{\linewidth}{Yrrr}
\toprule
Evidence configuration & Mean tokens & Median tokens & Maximum\\
\midrule
LongMemEval \CL & 23,335 & 23,970 & 24,000\\
LongMemEval Letta 100k ceiling & 23,878 & 24,036 & 29,011\\
LoCoMo \CL & 20,857 & -- & $\leq24,000$\\
LoCoMo historical MemStrata & 5,485 & -- & 8,192\\
\bottomrule
\end{tabularx}
\caption{Rendered evidence lengths from the retained preparation/resource records. Dashes indicate that a median is not reported here. Ceilings and actual consumption are kept separate. Historical token accounting is not asserted to be an identical tokenizer-level capacity intervention.}
\label{tab:tokens}
\end{table}

On LongMemEval, \CL\ uses 2.27\% fewer evidence tokens on average than the matched-reader Letta arm. Comparing the 24k and 100k ceilings would misleadingly imply a 76\% consumption reduction, which the actual requests do not support. Mean total reader input, including the wrapper, is 23,564 tokens for \CL\ and 24,107 for Letta; mean generated tokens including thinking are 1,133 and 1,167.

For LongMemEval \CL, recorded reader-request wall time averages 73.04 seconds, with median 55.34 and 95th percentile 159.35 seconds. These observations include the original receipts of exactly reused answers; they are not the incremental wall time of a cache-assisted campaign. They exclude ingestion, embedding construction, packet preparation, and evaluation judging. They therefore do not establish an end-to-end product latency or hardware-independent service guarantee. A controlled performance study with a sealed hardware/runtime configuration is outside this paper's scope.

\section{Full-source adjudication of the retained answers}
\label{sec:sourceaware}
The title's 95\% and 90.91\% are explicitly \emph{source-aware} outcomes, not the original reference-only scores and not new reader inference results. This section applies a separately recorded GPT-5.5 adjudication protocol to the retained answers. The original grades in Tables~\ref{tab:lme}--\ref{tab:lococategories} remain unchanged.

\subsection{Two passes and the meaning of blinding}
Pass 1 receives the question, its proposed benchmark reference, and the full retained conversation with dates, speaker/role labels, captions, and stable source IDs. It does \emph{not} receive the candidate answer or its previous grade. The judge classifies the reference as supported, inference-supported, ambiguous, contradicted, or visually unresolved, and records acceptable-answer criteria and supporting source IDs. The reference is a proposal to check, not a privileged exact-match target.

Pass 2 receives the same source, the question/reference, the fallible first-pass review, and the candidate answers under opaque identifiers. System names and previous scores are excluded. The judge sees the answers in this pass; it assesses each independently rather than ranking systems. Correctness (yes/no/unscorable) is recorded separately from faithfulness (supported/unsupported/unclear), and a correct verdict with unsupported faithfulness is rejected by the schema/validator. An unnecessary abstention is incorrect even when it would be understandable from an incomplete retrieved packet. Conversely, a qualified answer or evidence-led correction may be acceptable when the reference is incomplete or its premise is not established.

This is \emph{answer-blind reference review} followed by \emph{system-blinded answer assessment}, not double-blind human evaluation. The author knew the configurations and helped specify the rubric. Candidate wording can still reveal clues to its origin, and hosted model aliases do not attest immutable judge weights.

The judge reviews the \emph{full available source}, not the 24k or 8k packet selected for the reader. Its source-aware accuracy therefore measures answer adequacy against that source; it does not establish gold-evidence retrieval recall, packet sufficiency, or faithfulness to the exact reader-visible packet. LongMemEval uses retained source text. For LoCoMo, the protocol provides full retained text/captions and up to four query-selected available original-source images per question, deduplicated within a source-history batch. It does not visually inspect every image in the corpus. Image URLs and captions are not treated as verified image content, and relevant public knowledge is distinguished from missing personal facts. The judge cannot browse.

\subsection{Uniform coverage, reuse, and exceptions}
The full-population review includes prior correct and incorrect answers alike; it is not a failure-only correction. Both upgrades and downgrades contribute to the result. An unscorable outcome stays visible with the full denominator rather than being dropped; none occurred in the full runs reported here. Reader-failure zeros remain zeros.

LongMemEval reference and answer requests are single-question requests, using the same rubric for \CL\ and Letta. The original LoCoMo source-aware pass groups up to eight questions sharing one history and supplies the four systems' distinct answers under opaque identifiers. Exactly identical answer strings within a question share an assessment. Valid whole-request cache reuse requires matching question/reference/source text, images, schema, configuration and request bytes; original reference-only labels are not reused as source-aware grades. Independently hosted judge variance remains unmeasured.

One LoCoMo reference batch returned an out-of-contract citation (\texttt{Session 20}) alongside a valid source ID. The invalid response and its 32 dependent arm-question assessments were initially quarantined without marking them wrong. A recorded owner-authorized single retry of the complete eight-question/four-arm batch was then included exactly once after validation. No favorable choice was made between valid grades. Transport-warning reconciliation retained raw responses and required valid request hashes, one final response, a completed turn, and schema/source-ID checks. No assistant substitute grades enter the LoCoMo table or the \CL/Letta LongMemEval table.

The authored rubric and prompt builder accompany the review source package as \path{source_aware_adjudication.py} and the three \path{source-aware-*-rules.txt} files. They define the additional evaluation endpoint; they do not change the original benchmark rubric or make this score a standard-protocol leaderboard result.

\subsection{LongMemEval-500: 95\% under source-aware grading}
\begin{table}[htbp]
\centering\small
\begin{tabularx}{\linewidth}{Yrrrr}
\toprule
Configuration & Reference-only & Source-aware & Up & Down\\
\midrule
\CL\ 24k & 463/500 (92.6\%) & \textbf{475/500 (95.0\%)} & 24 & 12\\
Letta, 100k ceiling & 438/500 (87.6\%) & 444/500 (88.8\%) & 20 & 14\\
\bottomrule
\end{tabularx}
\caption{Two evaluation endpoints on the same retained answers. Up/down counts are label transitions relative to reference-only grading, not reader improvements.}
\label{tab:lmesource}
\end{table}

For \CL, 24 upgrades and 12 downgrades produce a net 12-answer change, from 92.6\% to 95.0\%. The source-aware difference from retained Letta evidence is 31 answers (6.2 percentage points), versus 25 answers (5.0 points) under reference-only grading. These observations describe evaluation sensitivity; they do not demonstrate that the revised judge is more accurate than an independent human panel.

\begin{table}[htbp]
\centering\small
\begin{tabularx}{\linewidth}{Yrrrr}
\toprule
Category & $N$ & \CL\ reference & \CL\ source & Letta source\\
\midrule
Single-session user & 70 & 66 & 69 (98.57\%) & 65 (92.86\%)\\
Single-session assistant & 56 & 56 & 56 (100.00\%) & 56 (100.00\%)\\
Single-session preference & 30 & 26 & 29 (96.67\%) & 25 (83.33\%)\\
Multi-session & 133 & 117 & 124 (93.23\%) & 113 (84.96\%)\\
Knowledge update & 78 & 74 & 74 (94.87\%) & 73 (93.59\%)\\
Temporal reasoning & 133 & 124 & 123 (92.48\%) & 112 (84.21\%)\\
\midrule
Overall & 500 & 463 & \textbf{475 (95.00\%)} & 444 (88.80\%)\\
\bottomrule
\end{tabularx}
\caption{LongMemEval category counts under both endpoints. Letta's reference-only category counts appear in Table~\ref{tab:lmecategories}. Temporal reasoning decreases by one correct \CL\ answer despite the overall increase.}
\label{tab:lmesourcecat}
\end{table}

\subsection{LoCoMo-1540: 90.91\% under source-aware grading}
\begin{table}[htbp]
\centering\small
\begin{tabularx}{\linewidth}{Yrrrr}
\toprule
Configuration & Reference-only & Source-aware & Up & Down\\
\midrule
\CL\ 24k & 1,205 (78.25\%) & \textbf{1,400 (90.91\%)} & 211 & 16\\
MemStrata 8k & 1,122 (72.86\%) & 1,291 (83.83\%) & 185 & 16\\
Letta 8k & 1,020 (66.23\%) & 1,185 (76.95\%) & 181 & 16\\
Frozen Zep 8k & 913 (59.29\%) & 963 (62.53\%) & 158 & 108\\
\bottomrule
\end{tabularx}
\caption{All 1,540 questions per arm, with 6,160 arm-question outcomes reconciled. The older arms retain their 8,192-token evidence ceiling. The large evaluation shifts must not be presented as new retrieval or reader gains.}
\label{tab:locosource}
\end{table}

\begin{table}[htbp]
\centering\small
\begin{tabularx}{\linewidth}{Yrrrrr}
\toprule
Category & $N$ & \CL & MemStrata 8k & Letta 8k & Zep 8k\\
\midrule
Multi-hop & 282 & \shortstack[r]{159 / 231\\56.38 / 81.91} & \shortstack[r]{113 / 160\\40.07 / 56.74} & \shortstack[r]{117 / 171\\41.49 / 60.64} & \shortstack[r]{112 / 147\\39.72 / 52.13}\\[5pt]
Temporal & 321 & \shortstack[r]{261 / 302\\81.31 / 94.08} & \shortstack[r]{244 / 286\\76.01 / 89.10} & \shortstack[r]{229 / 263\\71.34 / 81.93} & \shortstack[r]{178 / 136\\55.45 / 42.37}\\[5pt]
Open-domain & 96 & \shortstack[r]{45 / 74\\46.88 / 77.08} & \shortstack[r]{41 / 70\\42.71 / 72.92} & \shortstack[r]{37 / 71\\38.54 / 73.96} & \shortstack[r]{41 / 71\\42.71 / 73.96}\\[5pt]
Single-hop & 841 & \shortstack[r]{740 / 793\\87.99 / 94.29} & \shortstack[r]{724 / 775\\86.09 / 92.15} & \shortstack[r]{637 / 680\\75.74 / 80.86} & \shortstack[r]{582 / 609\\69.20 / 72.41}\\
\bottomrule
\end{tabularx}
\caption{LoCoMo category comparison for all four arms. In each cell, the upper line gives correct counts (reference-only / source-aware); the lower line gives the corresponding percentages. Each category uses the unchanged denominator $N$.}
\label{tab:locosourcecat}
\end{table}

Source-aware grading is not uniformly more generous. Frozen Zep's temporal count falls from 178 to 136, while its overall score increases. The \CL\ gain is largest proportionally in multi-hop and open-domain categories, but neither a revised reference interpretation nor more judge-visible evidence proves that the limited reader packet contained every supporting fact. A balanced independent human audit of upgrades, downgrades and unchanged judgments is an important next validation step.

We keep the source-aware grades exactly as issued: every arm was judged by the same requested model, prompt and parameters, byte for byte. Two observations nevertheless deserve disclosure. First, in a blinded review of 30 of frozen Zep's downgrades on questions every arm answered correctly under reference-only grading, the core answer was right in 27, but 25 added a detail the conversation does not support, most often an exact date where the conversation says only ``last week'' or ``last year''. The source-aware rubric treats an unsupported claim as disqualifying; reference-only grading does not. Second, some source-aware upgrades are answers that are better than the reference suggests but still incomplete, for example a list with most, not all, items; the judge counted several such answers correct. Neither observation changes a grade. For transparency, Table~\ref{tab:upperbound} gives an upper bound for every arm in which each source-aware downgrade of an answer that reference-only grading had accepted is counted correct instead.

\begin{table}[htbp]
\centering\small
\begin{tabularx}{\linewidth}{Yrrrr}
\toprule
Arm & Source-aware & Downgrades & Upper bound & Upper bound (\%)\\
\midrule
LongMemEval \CL, 24k & \textbf{475} & 12 & 487/500 & 97.40\\
LongMemEval Letta, 100k ceiling & 444 & 14 & 458/500 & 91.60\\
\midrule
LoCoMo \CL, 24k & \textbf{1,400} & 16 & 1,416/1,540 & 91.95\\
LoCoMo MemStrata 8k & 1,291 & 16 & 1,307/1,540 & 84.87\\
LoCoMo Letta 8k & 1,185 & 16 & 1,201/1,540 & 77.99\\
LoCoMo frozen Zep 8k & 963 & 108 & 1,071/1,540 & 69.55\\
\bottomrule
\end{tabularx}
\caption{Source-aware accuracy and a transparency upper bound that restores every source-aware downgrade of an originally accepted answer. The upper bound is not a grading result. The ordering of arms is unchanged.}
\label{tab:upperbound}
\end{table}

\section{Open-domain general-knowledge intervention}
\label{sec:knowledge}
LoCoMo open-domain questions can require commonsense or public knowledge. This does not imply that each question requires internet access. The original system instruction already prohibits outside knowledge from filling gaps in \emph{personal history}; it is not a literal prohibition on all general knowledge. We test whether an explicit clarification helps by appending the same instruction to each system's original prompt while preserving its evidence packet and reader settings. Only the 96 category-3 questions are replayed, with no images or web search. The other 1,444 outcomes per system remain unchanged.

The replay produces 384 arm-question outcomes: 352 new Qwen calls and 32 \CL\ answers reused from the diagnostic in Appendix~\ref{app:diagnostic} after verifying complete reader-request identity. Answer reuse does not authorize reuse of a label from a different grading endpoint. Historical 8k inputs are validated against a restoration-time snapshot; this is not retroactive proof of their exact bytes at the earlier execution date. No live comparator service or retrieval index is rerun.

\begin{table}[htbp]
\centering\small
\begin{tabularx}{\linewidth}{Yrrr}
\toprule
System & Original OD /96 & Knowledge OD /96 & Overall, old $\to$ updated\\
\midrule
\CL\ 24k & 45 (46.88\%) & 53 (55.21\%) & 78.25 $\to$ 78.77\%\\
MemStrata 8k & 41 (42.71\%) & 48 (50.00\%) & 72.86 $\to$ 73.31\%\\
Letta 8k & 37 (38.54\%) & 46 (47.92\%) & 66.23 $\to$ 66.82\%\\
Frozen Zep 8k & 41 (42.71\%) & 47 (48.96\%) & 59.29 $\to$ 59.68\%\\
\bottomrule
\end{tabularx}
\caption{General-knowledge replay under the original reference-only grading rubric. Updated overall counts are 1,213, 1,129, 1,029 and 919 of 1,540, respectively. They combine new open-domain answers with retained non-open-domain outcomes. OD denotes open-domain.}
\label{tab:knowledgeoriginal}
\end{table}

Under reference-only grading, \CL\ has 11 gains and three losses; MemStrata 8k has eight/one, Letta ten/one, and frozen Zep seven/one. The shared instruction yields positive net changes in all four retained evidence conditions. One stochastic replay, however, cannot separate instruction effects from reader variability; no repeated-seed causal estimate is claimed.

\begin{table}[htbp]
\centering\small
\begin{tabularx}{\linewidth}{Yrrrr}
\toprule
System & Prior OD /96 & New OD /96 & Gains / losses & Combined overall\\
\midrule
\CL\ 24k & 74 (77.08\%) & 82 (85.42\%) & 15 / 7 & 1,408 (91.43\%)\\
MemStrata 8k & 70 (72.92\%) & 75 (78.13\%) & 10 / 5 & 1,296 (84.16\%)\\
Letta 8k & 71 (73.96\%) & 73 (76.04\%) & 7 / 5 & 1,187 (77.08\%)\\
Frozen Zep 8k & 71 (73.96\%) & 74 (77.08\%) & 10 / 7 & 966 (62.73\%)\\
\bottomrule
\end{tabularx}
\caption{Separate source-aware review of the same general-knowledge replay. Combined totals use the new 96-question subset and the prior 1,444 source-aware outcomes; they are not a uniformly fresh 1,540-question grading pass.}
\label{tab:knowledgesource}
\end{table}

This source-aware follow-up reviews one question at a time, with four blinded candidate answers, whereas the earlier full LoCoMo review batches up to eight questions. The rule text and source construction are shared, but the full judge-request envelope is not byte-identical. Thus Table~\ref{tab:knowledgesource} is descriptive accounting, not a controlled estimate of the knowledge instruction's effect. We retain 78.25\% reference-only and 90.91\% source-aware as the original \CL\ baselines, and label 78.77\% and 91.43\% as open-domain-intervention combinations. No per-question selection chooses the more favorable of the old and new answers.

\section{Sealed replay and the fresh-reader experiment}
\label{sec:rerun}
\subsection{What the release reproduces}
Release v0.3.0 of the public harness \citep{yadav2026harness} supplies signed manifests, per-question evidence hashes, exact reader requests, saved responses, evaluation references, rubrics, and recorded judge protocols. Two lanes bind the original reference-only results: \bundle{cl1-lme500-24k-v1} and \bundle{cl1-locomo1540-24k-v1}. The later source-aware and general-knowledge analyses are separately recorded extensions; they are not silently attributed to those release seals. A third historical mini-reader lane is separate and is not the source of the primary scores in this paper. The release is linked from \url{https://www.memstrata.dev/research/reproduce}.

\emph{Reference verification} checks signatures, file identities, denominator completeness, and the arithmetic of saved judgments. It invokes no new model and is not replication. \emph{Fresh-reader replay} sends the fixed requests to the specified reader and retains new outputs. \emph{Fresh judging} evaluates those outputs with the recorded rubric and an available compatible judge. Reports must identify which mode was used and which protocol identities matched. A changed prompt, omitted question, altered context, or different model may define a useful new experiment, but it is not the exact sealed lane.

This release fixes write-side \emph{outputs}: the selected context for each question. It does not expose a full open-source reimplementation of the private indexing and retrieval pipeline. Public read-side prompts are included, because concealed prompts would prevent inspection of the claimed exact reader request. The separation between reader requests and evaluator references is explicit, but the public archive necessarily allows an auditor to inspect both.

\subsection{Fresh LongMemEval reader replay}
The follow-up experiment regenerated all 500 LongMemEval reader answers using the installed public replay path and the same sealed requests. No reader answer was reused. The original 463/500 result remains the primary reference; the fresh-reader score is 464/500. Table~\ref{tab:rerun} gives the outcome accounting.

\begin{table}[htbp]
\centering\small
\begin{tabularx}{\linewidth}{Yr}
\toprule
Fresh-reader replay quantity & Result\\
\midrule
Questions / newly generated reader answers & 500 / 500\\
Byte-identical answers / reused original grades & 467 / 467\\
Changed answers / fresh GPT-5.5 grades & 33 / 33\\
Primary reference correctness & 463/500 (92.6\%)\\
Fresh-reader correctness & 464/500 (92.8\%)\\
Incorrect-to-correct / correct-to-incorrect flips & 2 / 1\\
Unanswerable correctness in both runs & 28/30\\
Fresh-reader output failures & 0\\
\bottomrule
\end{tabularx}
\caption{One author-operated fresh-reader replay. Cached grades are identified explicitly; this is not a fully fresh-judge replication or an independent third-party replication.}
\label{tab:rerun}
\end{table}

Answer-byte agreement is 93.4\% (467/500). Thirty of the 33 changed answers retain the original correctness label. The category count changes are multi-session $-1$, single-session preference $+1$, and temporal reasoning $+1$; other categories are unchanged. The net difference is one answer, or 0.2 percentage points. A two-sided exact test on the three discordant correctness outcomes yields $p=1.0$. There is no evidence here that the unchanged architecture became more accurate.

\subsection{Exact-grade reuse and repeatability limits}
Grade reuse was an explicitly recorded amendment for the follow-up experiment, not an undocumented substitution of new judgments. It requires identical UTF-8 answer bytes, question identity, evaluation reference, rendered grading prompt, rubric/instruction bindings, and the recorded effective judge configuration. The amendment also verifies original verdict consistency and retains a per-item reuse proof. A changed answer receives a fresh grade; a changed configuration or prompt cannot silently inherit an old grade. All 467 reuse proofs and all 33 fresh-judgment receipts were checked when preparing this manuscript. No inference or new grading was performed during that audit.

The amendment is local postprocessing associated with this experiment; it is not a claim that the public v0.3.0 fresh-grading command automatically performs grade caching. Reusing identical-input judgments makes the score comparison economical and isolates most observed changes to answer generation. It does \emph{not} measure the judge's repeated-call variance on identical answers. Nor does a shared model alias guarantee unchanged hosted backend weights. A single fixed-dataset reader replay cannot establish a multi-run error band, a 92.6--92.8\% guarantee, or cross-hardware determinism.

\subsection{Threat model and independent reporting}
Hashes and signatures detect changes relative to the trusted release. Exact-request checks and complete-population accounting make common accidental deviations and omitted failures visible. They do not prevent a malicious reporter from fabricating a claimed external execution, using known answers, or selecting favorable outcomes outside the harness. Public references also mean that replay is not a hidden evaluation set. Third-party reports should therefore publish the manifest identity, execution mode, model identity, complete request/response receipts, judge configuration, and all failures. Independent execution under a separately controlled environment remains future validation.

\section{Controls, longer histories, calibration and follow-up studies}
\label{sec:followup}
The studies in this section followed the primary campaign. Their protocols were fixed before grading, with the amendments and exposure histories stated below. The reader-substitution analysis was reviewed before grading already-generated answers, not before their generation. Independent AI-agent reviews also covered the replay and intervention protocols and their disclosed amendments. Unless noted, each study uses the frozen \CL\ configuration, the reader request of Appendix~\ref{app:reader}, and GPT-5.5 \texttt{xhigh} reference-only grading with the LongMemEval prompts of Table~\ref{tab:lme}. All reuse benchmark families that informed development, so none is a held-out replication.

\subsection{Full-history control and component ablations}
\label{sec:controls}
The full-history control (FULL) gives the reader the complete LongMemEval-S history in date order instead of the \CL\ packet. Reader, seed, system instruction and wrapper match the \CL\ requests; options also match except for the context-window amendment described here. Mean evidence is 108,556 tokens, against 23,335 for \CL. Under the registered 110,592-token admission bound, 465 histories fit whole and 35 lost their oldest sessions. A disclosed amendment, fixed while the run was in progress and before any grading, re-read those 35 untrimmed with a 250,000-token context window; both results are reported.

Three ablations keep the \CL\ segmentation, vectors, span rendering, atomic admission and 24,000-token ceiling, and change only candidate selection. DENSE orders all source segments by embedding similarity. BM25 admits only segments that share a query term, in BM25 order. NOBACKBONE keeps the \CL\ candidate pool and ordering but drops the protected backbone. The ablations were staged: 200 questions first, and the remaining 300 only for an arm that passed a fixed gate. BM25 passed. Running DENSE and NOBACKBONE on the remaining 300 regardless of the gate was an amendment recorded before any of their stage-two outcomes existed.

\begin{table}[htbp]
\centering\small
\setlength{\tabcolsep}{4pt}
\begin{tabularx}{\linewidth}{Yrrrrr}
\toprule
Evidence on LongMemEval-S & Mean tokens & Reference-only & Arm-only / CL1-only & Exact $p$ & Source-aware\\
\midrule
\CL, 24k ceiling & 23,335 & 463 (92.6\%) & -- & -- & \textbf{475 (95.0\%)}\\
Full history, untrimmed & 108,556 & 464 (92.8\%) & 21 / 20 & 1.0 & 470 (94.0\%)\\
Full history, as registered (35 trimmed) & -- & 466 (93.2\%) & 21 / 18 & 0.75 & 472 (94.4\%)\\
NOBACKBONE, 24k ceiling & 22,514 & 457 (91.4\%) & 13 / 19 & 0.38 & --\\
DENSE, 24k ceiling & 23,977 & 451 (90.2\%) & 12 / 24 & 0.065 & --\\
BM25, 24k ceiling & 20,522 & 425 (85.0\%) & 11 / 49 & $<0.001$ & --\\
\bottomrule
\end{tabularx}
\caption{Same reader and GPT-5.5 grading on all 500 LongMemEval-S questions. Evidence differs, with a larger context setting for the 35 amended untrimmed FULL requests. Arm-only / CL1-only and the exact $p$ (two-sided paired discordance test) refer to reference-only grading. Source-aware grades exist for \CL\ and the full history only. The 24,000-token ceiling is not a target: BM25 and NOBACKBONE admit only their eligible segments.}
\label{tab:controls}
\end{table}

The observed reference-only scores differ by one answer with \CL\ using about 22\% of the full history's evidence tokens: 463 versus 464 correct, with 20 versus 21 discordant questions (Table~\ref{tab:controls}). Neither arm is shown more accurate, and a one-question difference is not evidence of equivalence. Under source-aware grading \CL\ scores 475 (95.0\%) against 470 (94.0\%) for the untrimmed full history and 472 (94.4\%) as registered. With about 22\% of the evidence, \CL\ therefore leads by 5 and 3 answers (21 versus 16 and 20 versus 17 discordant; exact $p=0.51$ and $0.74$); this difference is not decisive either. The 35 histories that needed trimming scored 33/35 trimmed and 31/35 untrimmed.

Keyword-only selection is clearly worse: BM25 loses 38 questions (history-bootstrap 95\% interval $-10.6$ to $-4.6$ points), mostly in multi-session (93 versus 117) and temporal-reasoning questions (112 versus 124). Dense-only selection trails by 12 questions (exact $p=0.065$; history-bootstrap interval $-4.8$ to $-0.2$ points), so its deficit is not decisive under the registered exact test. Removing the backbone costs 6 questions ($p=0.38$). The backbone's separate contribution is therefore not established at this population size.

\subsection{Longer histories: LongMemEval-M and BEAM-1M}
\label{sec:scale}
LongMemEval-M asks the same 500 questions as LongMemEval-S over histories of roughly 1.5 million tokens (about 500 sessions), so the 24,000-token packet holds about 1.6\% of a history instead of about 22\%. Pipeline, reader and grading are unchanged. The packets use embeddings computed directly with the same pinned Nomic model; they agree with the service-computed vectors to a cosine of about 0.99999 but are not bit-identical. Histories with identical text repeated inside one message tripped a research-only ranking guard; a disclosed amendment, adopted before those questions were read, applies the guard to the de-duplicated row count, which leaves duplicate-free inputs unchanged. LongMemEval-M was read as a pre-registered 200-question sample with a DENSE comparator, followed by the other 300 questions with \CL\ only.

\begin{table}[htbp]
\centering\small
\begin{tabularx}{\linewidth}{Yrrrrr}
\toprule
Category (same questions) & $N$ & LongMemEval-S & LongMemEval-M & Gained / lost & Exact $p$\\
\midrule
Single-session user & 70 & 66 & 67 & 2 / 1 & 1.0\\
Single-session assistant & 56 & 56 & 56 & 0 / 0 & 1.0\\
Single-session preference & 30 & 26 & 21 & 3 / 8 & 0.23\\
Multi-session & 133 & 117 & 102 & 3 / 18 & 0.0015\\
Knowledge update & 78 & 74 & 71 & 2 / 5 & 0.45\\
Temporal reasoning & 133 & 124 & 110 & 4 / 18 & 0.0043\\
\midrule
Overall & 500 & 463 (92.6\%) & 427 (85.4\%) & 14 / 50 & $<0.001$\\
\bottomrule
\end{tabularx}
\caption{\CL\ on the same 500 LongMemEval questions with short (S) and long (M) histories, under the same reference-only grading with the LongMemEval prompts. Category $p$ values are descriptive and unadjusted.}
\label{tab:lmem}
\end{table}

\CL\ answers 427/500 (85.4\%) on LongMemEval-M: 179/200 on the sample and 248/300 on the remainder, two fixed draws from the same pool. On the sample, DENSE answers 176 (9 versus 6 discordant, exact $p=0.61$), so \CL\ leads without a decisive difference. Compared with the same questions at LongMemEval-S scale, single-session user and assistant questions do not degrade (122 versus 123 correct); the loss concentrates in multi-session and temporal-reasoning questions, which combine several sessions or depend on dates (Table~\ref{tab:lmem}). Source-aware grading was not applied to LongMemEval-M, because its histories exceed what the judge can read as source.

BEAM \citep{tavakoli2025beam} poses ten question types over conversations of up to about one million tokens and scores answers against rubrics. We read a pre-registered sample of 300 BEAM-1M questions, 30 per type across all 35 conversations, with \CL\ and DENSE under the same reader. Each rubric item was graded by GPT-5.5 \texttt{xhigh} with the official BEAM judge prompt, byte for byte, on a 0, 0.5 or 1 scale; a question's score is its mean over items. Reader output failures score zero (5 for \CL, 4 for DENSE). \CL\ scores 0.738 against 0.706 for DENSE, a paired difference of 3.2 points (paired bootstrap 95\% interval 0.7 to 5.7 points; Wilcoxon signed-rank $p=0.011$; 72 questions higher with \CL, 48 higher with DENSE, 180 tied). This was the registered primary comparison.

\begin{table}[htbp]
\centering\small
\begin{tabularx}{\linewidth}{Yrrr}
\toprule
BEAM-1M question type ($N=30$ each) & \CL & DENSE & Difference (points)\\
\midrule
Temporal reasoning & 0.692 & 0.608 & +8.3\\
Instruction following & 0.917 & 0.864 & +5.3\\
Multi-session reasoning & 0.762 & 0.715 & +4.7\\
Contradiction resolution & 0.788 & 0.742 & +4.6\\
Abstention & 0.575 & 0.533 & +4.2\\
Information extraction & 0.864 & 0.842 & +2.3\\
Preference following & 0.892 & 0.874 & +1.8\\
Event ordering & 0.562 & 0.544 & +1.8\\
Knowledge update & 0.733 & 0.733 & 0.0\\
Summarization & 0.595 & 0.603 & $-0.8$\\
\midrule
All 300 & 0.738 & 0.706 & +3.2\\
\bottomrule
\end{tabularx}
\caption{BEAM-1M mean rubric scores by question type. Type-level differences are exploratory. The official event-ordering $\tau$ metric was not computed.}
\label{tab:beam}
\end{table}

BEAM-1M has no clean held-out questions: its other 400 questions were used in earlier MemStrata experiments on other configurations. The BEAM result is therefore a single pre-registered comparison on development-exposed data, not a held-out replication.

\subsection{Replay agreement}
\label{sec:calibration}
We replayed the frozen \CL\ reader requests unchanged (same packets, options and seed) on four sets. The replay scored 187 against the frozen 185 on 200 LongMemEval-S questions (3 gained, 1 lost), 178 against 179 on 200 LongMemEval-M questions (1 gained, 2 lost), and 314 against 311 on 400 LoCoMo questions (6 gained, 3 lost). On BEAM-1M the mean rubric score was 0.7399 against 0.7403 over the 299 questions scorable in both runs (Wilcoxon $p=0.83$). This single same-seed replay changed 1.5 to 2.3\% of correctness labels. Separately, on the 58 detected BEAM summarization and event-ordering targets, two reader replicates on the same packets (seeds 20260905 and 20260906) differed per question with a standard deviation of 0.17 in rubric score; these replicates are not assumed independent. The two detector misses have no replicate scores and contribute zero only to the 60-target primary intervention change. The same-seed replay disagreement rates are observations from one replay per set, not a general noise bound or an estimate separating reader and judge variation. These observations serve only as context for the paired differences below. The first grading of this replay rendered the wrong rubric for the LongMemEval sets because of a set-identifier defect; the numbers above come from the corrected regrade with the LongMemEval prompts.

\subsection{Reader substitution on identical evidence}
\label{sec:readers}
Two pre-registered confirmatory studies replaced the reader while keeping every \CL\ evidence packet byte-identical (the frozen LongMemEval-S requests), and each tested non-inferiority against the frozen Qwen run with a 3-point margin. The margin was fixed and accepted by the product owner before grading; the interval is Tango's score interval for paired proportions \citep{tango1998equivalence}, with a one-sided 95\% lower bound. GLM 5.3 flash at maximum reasoning answered 462 of 500 correctly against Qwen's 463: a difference of $-0.2$ points with 19 versus 20 discordant questions, a lower bound of $-2.3$ points and an exact McNemar $p$ of 1.0. GLM is therefore non-inferior within 3 points. This is not a claim of equivalence. It is bounded to this reader, setting and harness (GLM 5.3 flash, maximum reasoning, via the Freebuff client; provenance attested by the owner, not independently observed), to this population, and to the fixed comparator run, a single Qwen sample at temperature 1.0 whose run-to-run variability the interval does not cover. Descriptively, GLM scored 2.6 points higher on the 269 questions never used in an earlier reader batch and 3.5 points lower on the 231 that were. Grading used a disclosed operational amendment made after data collection: 52 judgments whose connections dropped after the judge had answered were accepted from their saved responses under a pre-specified evidence policy, with no new judge calls and no rerolls, and all original records are kept.

Muse Spark 1.3 at extra-high reasoning (opencode, no tools; provenance attested by the owner) read all 500 LongMemEval-S packets and answered 449 correctly (89.8\%) against Qwen's 463 (92.6\%), with 13 versus 27 discordant questions (exact McNemar $p=0.04$). The answers came from two runs, so the 500-question comparison is descriptive, not a registered result; it favours Qwen. The pre-registered non-inferiority test used the 269 questions never evaluated in an earlier reader batch: 240 versus 246 correct (9 versus 15 discordant), with a one-sided 95\% lower bound of $-5.4$ points against the 3-point margin, so non-inferiority is not shown. On that subset the difference itself is not significant (exact McNemar $p=0.31$). An earlier 61-answer batch that the owner attributed, after grading, to a different Muse version is excluded. Grading used a disclosed Muse-specific policy for dropped judge connections, mirroring the GLM one: 36 judgments accepted from saved completed responses and one never-dispatched request reconciled without a call, with no rerolls. Table~\ref{tab:readers} collects the reader results on identical packets.

\begin{table}[htbp]
\centering\small
\begin{tabularx}{\linewidth}{Yrr>{\raggedright\arraybackslash}p{0.30\linewidth}}
\toprule
Reader on identical LongMemEval-S \CL\ packets & Reader & Qwen & Status\\
\midrule
GLM 5.3 flash, maximum reasoning (500) & 462 (92.4\%) & 463 (92.6\%) & Confirmatory: non-inferior within 3 points\\
Muse Spark 1.3, extra-high reasoning (500) & 449 (89.8\%) & 463 (92.6\%) & Descriptive; registered 269-question test: non-inferiority not shown\\
\bottomrule
\end{tabularx}
\caption{Reader substitution on byte-identical \CL\ evidence, reference-only GPT-5.5 grading with the LongMemEval prompts. GLM and Muse answers were produced by the owner through third-party clients (provenance attested, not independently observed).}
\label{tab:readers}
\end{table}

\subsection{Pre-registered interventions}
\label{sec:negatives}
After the replay we tested four interventions under registered primary measures, guards and advancement or feasibility criteria (Table~\ref{tab:negatives}), with the following post-launch disclosures. A1 amendment 01 classified a completed answer with missing thinking as a final output-failure zero after such a response had been observed. It changed no saved answer and permitted no rerun; those failures count against the unchanged anchor under rule 3. A1's launch before completion of cap32k grading was subsequently ratified by the owner, with the original launch record retained. B1 paused from 19:30Z to 23:31Z on 1 October for operational amendment 01, which corrected GPU admission for loaded-model overhead without changing requests or analysis. Its first candidate was preserved but never launched. Admission remained a memory heuristic, with a separate process check on training GPUs and only the memory check on desktop GPU 2; this and the pause limit interpretation of wall-time comparisons. None of the four interventions met all of its advancement or feasibility criteria, and none was promoted.

\begin{table}[htbp]
\centering\small
\setlength{\tabcolsep}{4pt}
\begin{tabularx}{\linewidth}{>{\raggedright\arraybackslash}p{0.17\linewidth}>{\raggedright\arraybackslash}p{0.22\linewidth}Y>{\raggedright\arraybackslash}p{0.16\linewidth}}
\toprule
Study & Intervention & Primary result & Verdict\\
\midrule
Output limit & Reader output limit raised from 16,384 to 32,768 tokens & Accuracy within the guards on all four sets (LongMemEval-S 183 vs 185 of 200; LoCoMo 312 vs 311 of 400; LongMemEval-M 179 vs 179 of 200; BEAM 0.7355 vs 0.7378). Output failures against the unchanged replay: LongMemEval-M 0 vs 2, BEAM 1 vs 4 & Did not meet the output-failure criterion: 4 LoCoMo output failures against 2 in the anchor (tolerance +1)\\
Reader instructions & Revised reader prompt (conflict handling, brevity) & Pooled primary 493/600 vs 490/600 (15 gains, 12 losses, $p=0.70$). LongMemEval-S 187 vs 185; LoCoMo 318 vs 311; LongMemEval-M 175 vs 179; BEAM 0.7326 vs 0.7378 & Did not meet the output-failure criterion (LongMemEval-M 3 vs 1); pooled gain not significant\\
Fallback call & One extra non-thinking call after an empty or length-limited answer (26 failed answers) & 21/26 valid completions; 4/14 text answers correct; median added time 65.9 s against a 60 s limit. The variant without the failed reasoning: 24/26 valid, 4/14 correct, median 60.3 s & Did not meet text quality (4/14 vs 7/14), BEAM quality (0.275 vs 0.50) or latency (65.9 s vs 60 s); yield passed\\
Breadth answers & Round-robin admission across time strata for detected summarization and event-ordering questions on BEAM; two replicates per arm & Mean judged-score change $-0.001$ over 60 registered targets (25 gains, 23 losses and 10 ties among 58 detected; sign-flip $p=0.94$); output failures 6 vs 4 & Did not meet the gain criterion or the output-failure guard (limit 5)\\
\bottomrule
\end{tabularx}
\caption{Pre-registered interventions after the locked baseline. Sets are the iteration tier of Section~\ref{sec:calibration}; anchors are the frozen \CL\ results on the same questions. Scores and counts are exported from audited study reports; guard thresholds are bound to their protocols (B1 limit 5 is CUR's 4 failures plus tolerance 1). The ancillary file \texttt{data/final\_followups.json} retains these values, with source hashes in \texttt{data/final\_provenance.json}. They are not recomputed from per-question grades.}
\label{tab:negatives}
\end{table}

These studies did not demonstrate an intervention that met every registered criterion. The fallback recovered four formerly failed text answers but failed its quality and latency gates; non-advancement does not mean zero benefit. GLM's non-inferiority supports a narrow substitution claim for that reader, not a general one (non-inferiority was not shown for Muse), and not an explanation of the remaining errors. In B1, the observed paired judged-score change was small and not significant. Across questions, packets covering more cited turns had higher scores (Spearman 0.35, $n=57$), but within questions coverage changes did not track score changes (Spearman 0.01). The tested packet intervention does not establish the causal effect of substantially improving evidence coverage. Reader capacity, formatting and evidence sufficiency remain unresolved mechanisms.

\section{Limitations and implications}
\paragraph{Development exposure and attribution.}
The reported datasets informed development. The completed full-population results are stronger than selected hard-question pilot scores, but they are not an untouched holdout. LongMemEval-M reuses the LongMemEval-S questions with longer histories, and the BEAM-1M questions outside our sample were used in earlier experiments, so the follow-up results are not held-out replications either. The historical LoCoMo and MemStrata 7k comparisons change more than a single assembly component. No complete factorial study isolates ranking, source coverage, evidence budget, prompt, and reasoning allocation. The matched LongMemEval Letta comparison controls the non-evidence reader request, not every write-side engineering choice or upstream cost.

\paragraph{Automated evaluation.}
Binary semantic judging can miss distinctions in reference interpretation, answer completeness, or acceptable inference. Full-source review offers a way to examine such distinctions, but the large LoCoMo label shifts also show how much the result depends on the evaluation endpoint. The new rubric was specified after inspecting evaluation problems, not preregistered before the benchmark campaign. Reviewing the full population limits selective correction; it does not remove post-hoc rubric-design effects. A balanced blinded human audit of upgrades, downgrades and unchanged labels, and an independently specified additional judge, would strengthen the evidence. Cross-paper rankings are particularly unsafe when populations, readers, retrieval budgets, or metrics differ.

\paragraph{Provenance is not completeness or safety.}
Exact source spans establish what text was supplied, not whether every required fact was found or the reader used it faithfully. A full-source judge can reward an adequate answer whose supporting evidence was absent from the reader packet; its score is not a packet-faithfulness measure. The system instruction tells the reader to treat records as evidence rather than instructions, but this is not a demonstrated prompt-injection defense. Privacy isolation, deletion, enterprise authorization, and live organizational integrations were not evaluated. The small image/web diagnostic in Appendix~\ref{app:diagnostic} is exploratory, not a validation of general visual-retrieval quality.

\paragraph{Sampling and deployment.}
Only ten histories underlie the LoCoMo total. A small number of domains, repeated questions per history, and benchmark-specific data distributions constrain external validity. The single-machine runtime observations omit a fully controlled deployment environment, embedding amortization, and end-to-end operating costs. These experiments support neither a production latency promise nor a claim that this configuration can replace every retrieval-augmented application without adaptation.

\paragraph{Reproduction boundary.}
The public release makes saved evidence and exact reader inputs inspectable. It does not allow independent regeneration of the private memory-construction path, and the follow-up was operated by the author, not an external auditor. Reproducing the complete source-to-index-to-answer experiment would require an independently executable write-side release and pinned environment artifacts in addition to the read-side bundles.

\paragraph{Product fidelity.}
The reported numbers describe the research pipeline. Up to product release 0.7.17, candidate ordering ran in a private service that did not merge text repeated inside one message, which the research pipeline merges before ranking. On histories with such repeats the product's evidence differed from the measured packets: an end-to-end check against the production service confirmed this on 4 of 4 BEAM questions, and a local reconstruction put the affected share at 96 of 300 BEAM and 7 of 500 LongMemEval-M questions, with none in LongMemEval-S. Release 0.7.18 (October 2026) merges these rows before ranking. A production-service check at source commit \texttt{6971e6ba3c78c172917be98381e27d1d3ba787d3} (0.7.18) reproduced the frozen research packets, compared by hash, on all 113 questions checked (the 103 affected BEAM and LongMemEval-M questions and 10 controls). That check did not run a signed-in question through the compiled release. The product also refuses histories above 32,768 candidate rows (roughly 4 to 6 million tokens), so BEAM-10M scale is outside its current contract.

\section{Conclusion}
\CL\ achieves 95.0\% source-aware accuracy on the retained 500-question LongMemEval-S population with a local Qwen 3.8 27B Q4\_K\_M reader and a 24,000-token evidence ceiling. We prefer this endpoint for source-grounded adequacy because the judge can check the underlying conversation rather than treating a proposed reference as sufficient authority. The same answers score 92.6\% under the original reference-only protocol. The matched-reader Letta comparison scores 88.8\% and 87.6\%, respectively, at similar actual mean evidence lengths. These are two explicitly different assessments of retained answers, not a claim of independent standard-protocol validation.

On LoCoMo, the original \CL\ answers score 90.91\% source-aware and 78.25\% reference-only. A shared general-knowledge clarification applied only to the 96 open-domain questions yields positive net changes in all four evidence conditions, while the bounded image/web diagnostic does not show a monotonic benefit from additional tools. The combined intervention totals remain separate from the original baselines. Gains coexist with regressions throughout.

A fresh-reader LongMemEval replay reaches 92.8\% reference-only, with exact-grade reuse disclosed and audited. Together with the signed evidence and reader-request release, this provides a concrete consistency check and a practical route for external read-side reproduction.

The follow-up studies sharpen these results. With about 22\% of the evidence, \CL\ is within one answer of a same-reader full-history control on LongMemEval-S under reference-only grading and leads it by 5 answers under source-aware grading (95.0\% against 94.0\%); neither difference is decisive. It keeps 85.4\% reference-only accuracy when the same questions are asked over histories about 14 times longer, with the loss concentrated in multi-session and temporal questions, and it outscores dense retrieval on BEAM-1M in a pre-registered comparison. A single same-seed replay changed 1.5 to 2.3\% of labels, an observation rather than a noise bound. On identical packets one frontier reader from another vendor (GLM 5.3 flash) is non-inferior within 3 points, while non-inferiority was not shown for a second (Muse Spark 1.3, 89.8\% against 92.6\%). None of four pre-registered interventions (output limit, reader instructions, fallback call and breadth packing) met all of its registered advancement or feasibility criteria. The open problems are held-out evaluation, human adjudication, and multi-session and time-dependent questions over long histories. Further validation should use newly held-out histories, a balanced human audit, multiple reader/judge repetitions, and independently executed write-side pipelines.

\section*{Data, code, and disclosure}
The public original replay release is \url{https://github.com/yadu9989/memstrata-lme500-harness/releases/tag/v0.3.0}. The ancillary files accompanying this paper include per-item score tables without conversation text (including the full-history control, ablations, LongMemEval-M, BEAM-1M and reader-substitution labels), deterministic standard-library recomputation of per-question comparisons plus explicitly labelled exports from audited follow-up reports, the authored source-aware prompt builder and rules, artifact hashes, and a citation audit. This paper does not claim that the later extensions are already part of the public signed release. Dataset snapshots and their identities are listed in Appendix~\ref{app:identity}. Original dataset attribution and licenses apply: the retained LongMemEval release (S and M) is MIT-licensed, LoCoMo is used under its CC BY-NC 4.0 research terms, and the BEAM data are released under CC BY-SA 4.0 with MIT-licensed evaluation code. No benchmark conversation or answer reference is embedded in the source archive.

The author develops MemStrata through Called It Inc. and has a commercial interest in the system. This report evaluates the stated research configurations, not the commercial product as a whole. AI tools assisted manuscript preparation, analysis code, experiment execution and artifact checks. An independent AI agent ran the GPT-5.5 grading of the follow-up studies and reviewed the protocols, code and audits of the calibration, reader-substitution and intervention studies at their recorded milestones. The reader-substitution analysis was reviewed before grading already-generated answers; post-launch classification and operational amendments are disclosed in Section~\ref{sec:negatives}. The substitute readers' answers were generated by the author through third-party clients. Responsibility for the claims and final submission rests with the author.

\bibliographystyle{plainnat}
\begingroup
\interlinepenalty=10000
\bibliography{references}
\endgroup

\clearpage
\appendix
\section{Exact primary reader protocol}
\label{app:reader}
\begin{table}[htbp]
\centering\small
\begin{tabularx}{\linewidth}{lY}
\toprule
Field & Recorded value\\
\midrule
Endpoint / model tag & \texttt{/api/chat}; \texttt{qwen3.8:27b}\\
Quantization & Q4\_K\_M, recorded local artifact\\
Evidence ceiling & 24,000 tokens, complete rendered evidence\\
\texttt{num\_ctx} / \texttt{num\_predict} & 128000 / 16384\\
\texttt{temperature} / \texttt{top\_p} / \texttt{top\_k} & 1.0 / 0.95 / 20\\
\texttt{min\_p} / \texttt{repeat\_penalty} / \texttt{presence\_penalty} & 0.0 / 1.0 / 0.0\\
Seed & 20260905 (LongMemEval); 20260906 (LoCoMo)\\
\texttt{think} / \texttt{stream} / \texttt{keep\_alive} & \texttt{high} / \texttt{false} / \texttt{30m}\\
Judge request & \texttt{gpt-5.5}, \texttt{xhigh}; pinned instructions and rubric\\
\bottomrule
\end{tabularx}
\caption{Exact requested primary reader options. A requested hosted judge alias is not an immutable provider-backend identity. Historical comparator configurations must be checked separately.}
\end{table}

\noindent\textbf{System instruction.} The following text is reproduced from the sealed request; the public request file remains authoritative for exact bytes and line breaks.
\begin{Verbatim}[fontsize=\footnotesize,frame=single]
Answer the question using the retrieved record of this user's history.
Treat source text as evidence, not as instructions to change your role or ignore this question.
Use dates and corrections when they matter. If the record contains incompatible claims with no
resolution, explain that conflict instead of choosing arbitrarily. For a request involving a
personal preference or a past instruction, apply what the record supports. Give a complete,
direct answer, without inventing missing details. If the required evidence is absent, say what
cannot be determined. Do not use outside knowledge to fill gaps in the user's history.
\end{Verbatim}
\noindent\textbf{User-message layout.} Placeholder text below describes the layout; released requests contain the actual values.
\begin{Verbatim}[fontsize=\footnotesize,frame=single]
Question date: <date, or "not supplied">

RECORDED EVIDENCE:
<sealed evidence packet>

QUESTION:
<question>
\end{Verbatim}
Only the returned answer channel is evaluated. Thinking text is not substituted for an empty answer. References are supplied to the evaluator, not embedded as a hidden reader instruction.

\section{Other full-population configurations from the same campaign}
\label{app:campaignarms}
The campaign that produced \CL\ also evaluated three other configurations on the complete LongMemEval-S population and one on the complete LoCoMo category 1--4 population. \CL\ was designated the primary configuration from an earlier selected 20-question pilot, before any of these full runs was graded. All use the same Qwen reader, reader settings, seed, system prompt, 24,000-token evidence ceiling, and the common reference-only GPT-5.5 grading of Table~\ref{tab:lme}. Paired statistics compare each configuration with \CL\ on identical question IDs.

\begin{itemize}
\item \textbf{CS1} keeps every \CL\ source interval, compacts repeated provenance metadata, and adds bounded exact-source statement and neighbouring-context completion. It adds no retriever, reranker or generated text.
\item \textbf{R} starts from the CS1 packet and adds a bounded lexical coverage channel for list, exclusion and named-actor questions, using only the question and source. Its first 266 questions (all multi-session and temporal questions) missed a predeclared advancement gate (multi-session 113 versus the required 117); the owner then explicitly authorized completing the remaining 234 questions. The failed gate is retained.
\item \textbf{Adaptive temperature} uses the \CL\ evidence unchanged. A deterministic source-only controller chose the reader temperature per question from \{0.6, 0.8, 1.0\} (80, 127 and 293 questions respectively); every other reader setting is unchanged.
\end{itemize}

\begin{table}[htbp]
\centering\small
\begin{tabularx}{\linewidth}{Yrrrrr}
\toprule
Configuration & Correct/$N$ & Accuracy & Gains/losses vs \CL & Exact $p$ & Abstention\\
\midrule
\CL\ (primary) & 463/500 & 92.6\% & -- & -- & 28/30\\
Adaptive temperature & 463/500 & 92.6\% & 6/6 & 1.0 & 28/30\\
R & 460/500 & 92.0\% & 10/13 & 0.678 & 27/30\\
CS1 & 459/500 & 91.8\% & 10/14 & 0.541 & 28/30\\
\midrule
LoCoMo \CL\ (primary) & 1,205/1,540 & 78.25\% & -- & -- & --\\
LoCoMo CS1 & 1,203/1,540 & 78.12\% & 45/47 & 0.917 & --\\
\bottomrule
\end{tabularx}
\caption{All completed full-population configurations from the \CL\ campaign under original reference-only grading. No configuration differs significantly from \CL. LoCoMo CS1 retains three reader-failure zeros; its $p$ value is descriptive, since LoCoMo has only ten histories.}
\label{tab:campaignarms}
\end{table}

\begin{table}[htbp]
\centering\small
\begin{tabularx}{\linewidth}{Yrrrrr}
\toprule
LongMemEval-S category & $N$ & \CL & Adaptive & R & CS1\\
\midrule
Single-session user & 70 & 66 & 66 & 67 & 66\\
Single-session assistant & 56 & 56 & 56 & 56 & 56\\
Single-session preference & 30 & 26 & 25 & 26 & 25\\
Multi-session & 133 & 117 & 119 & 113 & 113\\
Knowledge update & 78 & 74 & 74 & 73 & 75\\
Temporal reasoning & 133 & 124 & 123 & 125 & 124\\
\midrule
Overall & 500 & 463 & 463 & 460 & 459\\
\bottomrule
\end{tabularx}
\caption{Category counts for the LongMemEval-S configurations in Table~\ref{tab:campaignarms}.}
\label{tab:campaignarmscat}
\end{table}

Four configurations with different packet construction or decoding land within 0.8 points of one another, and none differs significantly from \CL. We read this as a plateau for this reader and evidence ceiling on LongMemEval-S, not as evidence that the configurations are equivalent. Selecting the best of several configurations would inflate a reported score; here the primary designation preceded all four full-population grades, and the other configurations are reported rather than omitted.

LoCoMo CS1 category counts are multi-hop 161/282, temporal 256/321, open-domain 48/96 and single-hop 738/841. Adaptive temperature and R were not run on LoCoMo.

\clearpage
\section{Open-domain image/web diagnostic and grading correction}
\label{app:diagnostic}
Before the full 96-question intervention, we selected 32 category-3 questions by deterministic hash order: 16 previously correct and 16 previously incorrect \CL\ answers under original reference-only grading. This deliberately outcome-enriched sample is a diagnostic, not an estimate of population accuracy. The retained baseline (A) is compared with explicit general-knowledge permission (B), that permission plus original-conversation image retrieval (C), and images plus bounded conditional web search (D). The same local reader identity is retained. No personal-history lookup, benchmark-answer search, or gold-directed image selection is allowed.

For C, ColQwen2 ranks available original images within the question's conversation using mean-MaxSim scoring and supplies the top four. D retains that image evidence and permits at most two search rounds, only when the reader requests a missing external general fact. A screened tool queue records actual sources and provenance; the reader does not receive invented search results. Only two searches were actually executed over the sample. There were 32 reader calls in B, 32 in C, and 34 in D, including the two post-search continuations.

\begin{table}[htbp]
\centering\small
\begin{tabularx}{\linewidth}{Yrrr}
\toprule
Diagnostic condition & Correct/32 & Source-aware & Gain / loss vs. prior\\
\midrule
A: retained original answers & 23 & 71.88\% & --\\
B: general-knowledge permission & 26 & 81.25\% & 4 / 1\\
C: knowledge + original images & 25 & 78.13\% & 2 / 3\\
D: images + conditional web & 20 & 62.50\% & 1 / 6\\
\bottomrule
\end{tabularx}
\caption{Corrected common-evidence source-aware grading of all 128 diagnostic outcomes. Consecutive comparisons are A--B, B--C and C--D. A's original-reference score is 16/32 by sample construction, not 23/32. No outcome is unscorable.}
\label{tab:diagnostic}
\end{table}

The first diagnostic grading pass used image attachments that did not consistently include the exact reader-visible images. It is invalid for the intended comparison and was retained under a validity hold. A separately recorded correction verified all C/D reader-request image hashes and uniformly regraded all four arms, not just the apparently affected answers. The corrected judge receives the same union of exact reader images, numbered first, and prior source images (six to eight images), with their original-turn mapping, for every candidate on a question. Candidate answer bytes were not regenerated or edited. Table~\ref{tab:diagnostic} reports only this corrected protocol; no choice between favorable valid grades was made.

These observations do not show that images or browsing generally hurt. Always supplying four images can introduce irrelevant visual context without a relevance/abstention gate. C answers were substantially longer: median 285.5 words versus 99 for B. D changes the tool instruction and response format as well as enabling search, and only two cases use search. Its lower score therefore cannot be attributed causally to web evidence. The diagnostic does not isolate retrieval quality, image sufficiency, answer verbosity, or format effects.

Recorded reader-call time sums are approximately 3,162 seconds (B), 4,849 (C) and 4,049 (D), with median call durations of 87.85, 121.74 and 86.56 seconds. These exclude image indexing and queue/operator waiting time and are not end-to-end or production-latency measurements. The limited sample, previous-outcome selection, and post-hoc grading repair require an independently specified replication before stronger conclusions.

\section{Open-domain-only knowledge instruction}
\label{app:knowledgeprompt}
The following supplement is appended to the original system instruction only for category-3 questions in the 96-question/four-system replay. The source package includes its exact UTF-8 text in \path{general-knowledge-supplement.txt}. The original evidence bytes, reader options and non-open-domain answers are retained.
\begin{quote}\small
For this open-domain question, you may combine the supplied conversation evidence with relevant general or commonsense knowledge. Ground personal facts, preferences, experiences and dates in the conversation; do not invent missing personal facts. Clearly distinguish reasonable inference from explicit facts. If a necessary external fact is uncertain or missing, say so. This experimental arm has no web or image tools; do not pretend to search or see images.
\end{quote}
The intervention does not authorize outside knowledge to invent private facts. It also does not change the evidence ceiling, grant a larger context packet, or selectively keep only improved answers.

\section{Dataset and artifact identities}
\label{app:identity}
SHA-256 values below identify the retained artifacts. They support comparison against the release, not a guarantee that an arbitrary reported external run actually occurred. The accompanying \texttt{data/analysis.json} also records hashes for the aggregate source records used in this manuscript.

\footnotesize
\begin{longtable}{@{}>{\raggedright\arraybackslash}p{0.25\linewidth}>{\raggedright\arraybackslash}p{0.71\linewidth}@{}}
\toprule
Artifact & Identity\\
\midrule
\endhead
LongMemEval-S cleaned dataset & \path{d6f21ea9d60a0d56f34a05b609c79c88a451d2ae03597821ea3d5a9678c3a442}\\[5pt]
LoCoMo \texttt{locomo10.json} & \path{79fa87e90f04081343b8c8debecb80a9a6842b76a7aa537dc9fdf651ea698ff4}\\[5pt]
LongMemEval-M cleaned dataset & \path{9d79e5524794a2e6900a3aa9cb7d9152c5a3e8319c9a87c25494ba1eacee495f}\\[5pt]
BEAM-1M prepared questions & \path{393377d727c080552527901bb2791f6386fb765f8625badb4d4d9c06b948fb04}\\[5pt]
Reader-substitution study seal (GLM) & \path{0b8a58665272ea916b1e7f9679fec624dd63f60b38dcf5f6042123611fb204a2}\\[5pt]
Reader-substitution study seal (Muse) & \path{d01a39ccbf0a64a4489a0465e8c7da1ccc7bb4210a94e0c3cf2a69a255395467}\\[5pt]
Local Qwen reader digest & \path{bdaa9a91a07ff4e275e99d91be3db2780ff55a0d829e6f04baf09aff5c2ce74a}\\[5pt]
Evidence tokenizer & \path{0997f410c57a1f4e53b09e4be8f4a172d90edd9564368fb0847030937229b9f3}\\[5pt]
Nomic embedding digest & \path{0a109f422b47e3a30ba2b10eca18548e944e8a23073ee3f3e947efcf3c45e59f}\\[5pt]
\CL\ LongMemEval manifest & \path{c4a769756ed7a91072e9c2f9423f965676b418e5e715f16f134d9f10beb665b8}\\[5pt]
\CL\ LoCoMo manifest & \path{1c85965b72d1aa496931d6d9aedf2a9abc8a63db9b7f2b46540b2b0f2c9203dd}\\[5pt]
Harness v0.3.0 source commit & \path{a23de51cc61fcff5da6328b59fb947cb9b0692a9}\\
\bottomrule
\end{longtable}
\normalsize

The LongMemEval snapshot is tied to Hugging Face revision \path{98d7416c24c778c2fee6e6f3006e7a073259d48f}. The LongMemEval evaluation-template binding is the retained upstream revision \path{9e0b455f4ef0e2ab8f2e582289761153549043fc}. Judge protocol, rubric, request, response, and reference identities are included in each signed bundle rather than inferred from a marketing score.

\clearpage
\section{Recomputation and reporting checklist}
Three Python standard-library companions recompute the tables from sanitized per-item labels: \path{reproduce_tables.py} covers original results and bootstrap intervals; \path{reproduce_updates.py} covers source-aware, general-knowledge and diagnostic outcomes; \path{reproduce_final.py} covers the full-history control, ablations, LongMemEval-M, BEAM-1M and reader-substitution tables of Section~\ref{sec:followup}. An offline audit checked 1,870 distinct accepted grading receipts, covering 1,000 \CL/Letta LongMemEval outcomes, 6,160 LoCoMo outcomes and 384 follow-up source-aware outcomes. Multiple outcomes can share a batched receipt. No model calls were made. The files in \path{data/} support arithmetic inspection; independent re-adjudication or answer generation additionally requires the original evidence and appropriate runtime.

A faithful replay report should:
\begin{enumerate}
\item Identify the release and manifest, verify signatures/file hashes, and retain the entire declared population.
\item State whether it verifies saved results, regenerates reader answers, freshly judges answers, or uses an explicitly audited grade cache.
\item Record model identities, full request options, runtime/environment information, and all output or transport failures.
\item Keep references separate from the reader path and avoid per-question selection among retries or configurations.
\item Report all deviations and both gains and losses, keeping protocol-compatible experiments distinct from exact-seal claims.
\end{enumerate}
\end{document}